\documentclass[lettersize,journal]{IEEEtran}
\usepackage{amsmath,amsfonts}
\usepackage[ruled,vlined]{algorithm2e}
\usepackage{xcolor}

\definecolor{beige}{RGB}{255,229,180}   
\usepackage{soul}     % 提供高亮命令
\usepackage{amssymb}
\usepackage{algorithmic}
\usepackage{array}
\usepackage{textcomp}
\usepackage{stfloats}
\usepackage{url}
\usepackage{verbatim}
\usepackage{graphicx}
\usepackage{cite}
\usepackage{hyperref} 

\usepackage{float}
\usepackage{times}
\usepackage{mathptmx}
\usepackage{booktabs}
\usepackage{pifont}

\usepackage{subfigure}

\hypersetup{
    colorlinks=true,        % 启用彩色链接
    linkcolor=red,          % 设置内部链接（如目录、图片、表格）为红色
    filecolor=magenta,      % 设置文件链接颜色
    urlcolor=blue           % 设置 URL 链接颜色
}

\begin{document}

\title{LASFNet: A Lightweight Attention-Guided Self-Modulation Feature Fusion Network for Multimodal Object Detection
}

\author{Lei Hao, Lina Xu, Chang Liu, and Yanni Dong, \IEEEmembership{Senior Member,~IEEE}

        % <-this % stops a space

\thanks{This study was jointly supported by the National Natural Science Foundation of China under Grants 62222116 and 62171417. The numerical calculations in this paper have been done on the supercomputing system in the Supercomputing Center of Wuhan University. (\emph{Corresponding author: Yanni Dong})}

\thanks{Lei Hao is with the School of Geophysics and Geomatics, China University of Geosciences, Wuhan 430074, China, and also with the School of Resource and Environmental Sciences, Wuhan University, Wuhan 430079, China (e-mail: haolei@cug.edu.cn).}
\thanks{Lina Xu is with the School of Geophysics and Geomatics, China University of Geosciences, Wuhan 430074, China (e-mail: xulina@cug.edu.cn).}
\thanks{Chang Liu is with the School of Resource and Environmental Sciences, Wuhan University, Wuhan 430079, China (e-mail: liu\_chang\_@whu.edu.cn).}
\thanks{Yanni Dong is with the School of Resource and Environmental Sciences, Wuhan University, Wuhan 430079, China (e-mail:
dongyanni@whu.edu.cn).}
}

\markboth{IEEE Transactions on Cybernetics}%
{Hao \MakeLowercase{\textit{et al.}}: LASFNet:A Lightweight Attention-Guided Self-Modulation Feature Fusion Network for Multimodal Object Detection}

\IEEEpubid{}

\maketitle

\begin{abstract}
Effective deep feature extraction via feature-level fusion is crucial for multimodal object detection. However, previous studies often involve complex training processes that integrate modality-specific features by stacking multiple feature-level fusion units, leading to significant computational overhead. To address this issue, we propose a new fusion detection baseline that uses a single feature-level fusion unit to enable high-performance detection, thereby simplifying the training process. Based on this approach, we propose a lightweight attention-guided self-modulation feature fusion network (LASFNet), which introduces a novel attention-guided self-modulation feature fusion (ASFF) module that adaptively adjusts the responses of fusion features at both global and local levels based on attention information from different modalities, thereby promoting comprehensive and enriched feature generation. Additionally, a lightweight feature attention transformation module (FATM) is designed at the neck of LASFNet to enhance the focus on fused features and minimize information loss. Extensive experiments on three representative datasets demonstrate that, compared to state-of-the-art methods, our approach achieves a favorable efficiency-accuracy trade-off, reducing the number of parameters and computational cost by as much as 90\% and 85\%, respectively, while improving detection accuracy (mAP) by 1\%-3\%. The code will be open-sourced at https://github.com/leileilei2000/LASFNet.
\end{abstract}


\section{Introduction}
\IEEEPARstart{R}{ecently,} significant advancements\cite{zou2023object} have been made in object detection technologies based on visible light or infrared imaging\cite{gundougan2023ir},\cite{yin2024camoformer},\cite{Liu2024cdr},\cite{Yuan2024re}. This progress has enabled substantial applications in fields such as remote sensing\cite{zhang2024learning},\cite{Qin2024}, autonomous driving\cite{uzair2024channel},\cite{Zhu2025},medical diagnosis\cite{xiao2024ctnet},\cite{Dong2025}, traffic management\cite{liu2024tiny}, and industrial production\cite{Zuo2024mf}. However, the performance of visible light object detection is adversely affected by inadequate lighting conditions and occlusion of the object. Similarly, infrared object detection suffers from issues related to thermal source interference and lack of color information. To address these challenges, researchers have taken advantage of the complementary nature of infrared and visible light modalities, integrating information from both to implement multimodal object detection for improving the limitations of single-modality detection, with a positive response and rapid development\cite{zhang2023visible}. 

In multimodal object detection tasks, researchers typically employ data fusion techniques to combine information from different sensors, such as visible light and infrared modalities, to overcome the limitations of single-modal data. Based on the stage of fusion, existing multimodal object detection methods can be generally categorized into three types: pixel-level fusion\cite{Cao2022}, feature-level fusion\cite{zhao2023metafusion}, and decision-level fusion\cite{wang2024rgb}.  Pixel-level fusion aims to combine image data from different modalities at the pixel level, producing richer and higher-quality fused information to enhance detection accuracy. Decision-level fusion improves accuracy by independently detecting images from various sensors and then combining the results. In contrast, feature-level fusion integrates features from different modalities during the feature extraction phase to generate a comprehensive representation, allowing for a deeper exploration of the differences and relationships between modalities, thus achieving improved detection performance. Therefore, feature-level fusion has found broader applications in the field of multimodal object detection\cite{guo2025damsdet}. However, previous studies often overlook the significant computational overhead associated with feature-level fusion\mbox{\cite{yuan2022translation}}, leading to issues such as high power consumption and inference latency. These challenges may hinder the widespread adoption of multimodal detection in embedded systems (e.g., autonomous driving controllers, industrial robots) and resource-constrained edge devices (e.g., smart cameras, drones).

Recently, self-modulation mechanisms have demonstrated remarkable performance in multimodal object detection, as they can adaptively regulate the contribution of each modality's information, thereby enabling dynamic fusion of information across different modalities to enhance detection accuracy. For example, Yuan et al.\cite{yuan2024c} proposed a calibration-complementary mechanism that adaptively adjusts the complementary features between different modalities, improving detection accuracy. Xu et al.\cite{Xu2024air} improved small object detection by adaptively aggregating multimodal features and weighting key information. Dong et al.\cite{dong2024fusion} and Wang et al.\cite{wang2024mask} achieved deep fusion by adaptively adjusting modality differences in the hidden state space, thereby enhancing detection performance. Zhou et al.\cite{zhou2024dmm} adaptively addressed modality conflicts by capturing multimodal disparity information, resulting in efficient multimodal object detection. Although these methods have shown promising detection results through self-modulation mechanisms, the substantial computational cost of the models remains a significant concern. Therefore, a key challenge lies in developing low-power feature self-modulation techniques to facilitate the lightweight evolution of multimodal object detection in real-world applications.

\begin{figure*}[ht]
\centering
\includegraphics[width=\linewidth]{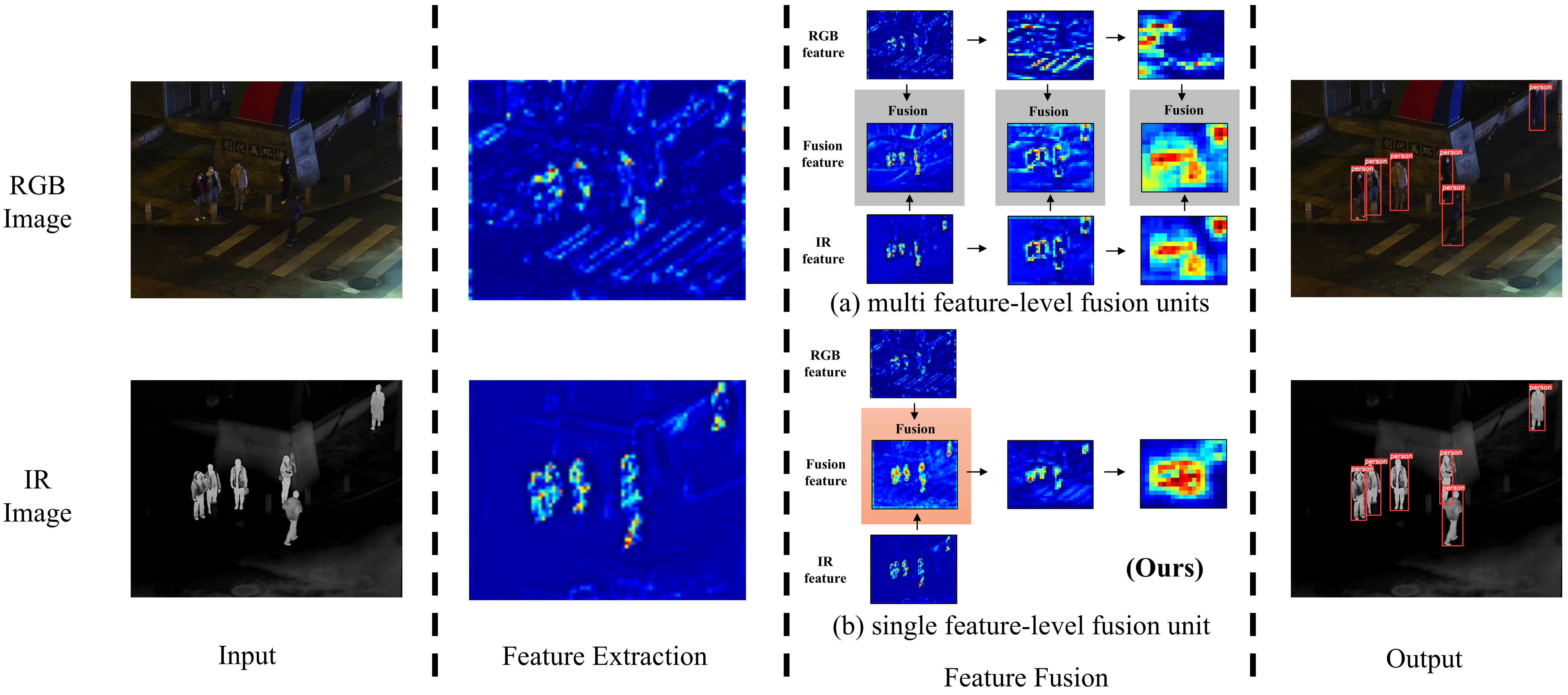}
\caption{Popular multimodal fusion baseline and the proposed new baseline. (a) Popular baseline with multi-feature fusion units: This approach requires a separate fusion of texture and semantic features from different modalities, which incurs substantial computational overhead to improve detection performance. (b) Proposed new baseline with single-feature fusion units: This method prioritizes the fusion of texture features, followed by the exploration of deeper semantic information within the fused features. Compared to the popular baseline, the proposed baseline achieves equally precise detection performance with a more lightweight and efficient design.  }
\label{fig1}
\end{figure*}
\setlength{\parskip}{0pt}

Towards this end, we propose a lightweight attention-guided self-modulation feature fusion network (LASFNet) for multimodal object detection, which achieves high-performance inference with low computational cost. Unlike the fusion strategies in previously popular multimodal detection baselines (as shown in Figure \ref{fig1}(a)), the proposed new baseline achieves lightweight and efficient detection by using a single feature-level fusion unit (as shown in Figure \ref{fig1}(b)). To obtain rich and robust fused features, we introduce a novel attention-guided self-modulation feature fusion (ASFF) module. ASFF performs self-modulation of global and local information under the guidance of attention to facilitate deep fusion. Furthermore, we develop a lightweight and flexible feature attention transformation module (FATM) to enhance the focus on the fused features in the deep layers of the network, effectively reducing information loss. We conduct extensive experiments on three representative datasets: DroneVehicle\cite{sun2022drone}, LLVIP\cite{jia2021llvip}, and VTUAV\_det\cite{zhang2023drone}. Experimental results demonstrate that LASFNet achieves an optimal accuracy-efficiency trade-off with a lightweight model compared to existing methods. The main contributions of this paper can be summarized as follows:

1) We propose a lightweight attention-guided self-modulation feature fusion network (LASFNet) for multimodal object detection. LASFNet follows the strategy of using a single feature fusion unit, achieving a lightweight model while maintaining accurate detection performance. 

2) We design an attention-guided self-modulation feature fusion (ASFF) module, which utilizes attention information from different modalities to perform adaptive deep fusion of global and local features based on their response levels. This module generates rich and robust representations by gradually exploring the information in the fused features from coarse to fine.

3) We propose a Feature Attention Transformation Module (FATM) at the neck of the network, which emphasizes the key representations of multimodal fused features to reduce redundant information interference and improve multi-scale feature fusion, thereby enhancing the detection capability for objects of different scales. 

The structure of this paper is as follows: In Section \ref{sec:related_work}, we briefly review related work. In Section \ref{sec:method}, we detail the LASFNet architecture. Section \ref{sec:experiments} presents comprehensive experimental results. Section \ref{sec:conclusions} provides a brief conclusion.
\begin{figure*}[!h]
 
\centering
\includegraphics[width=\textwidth]{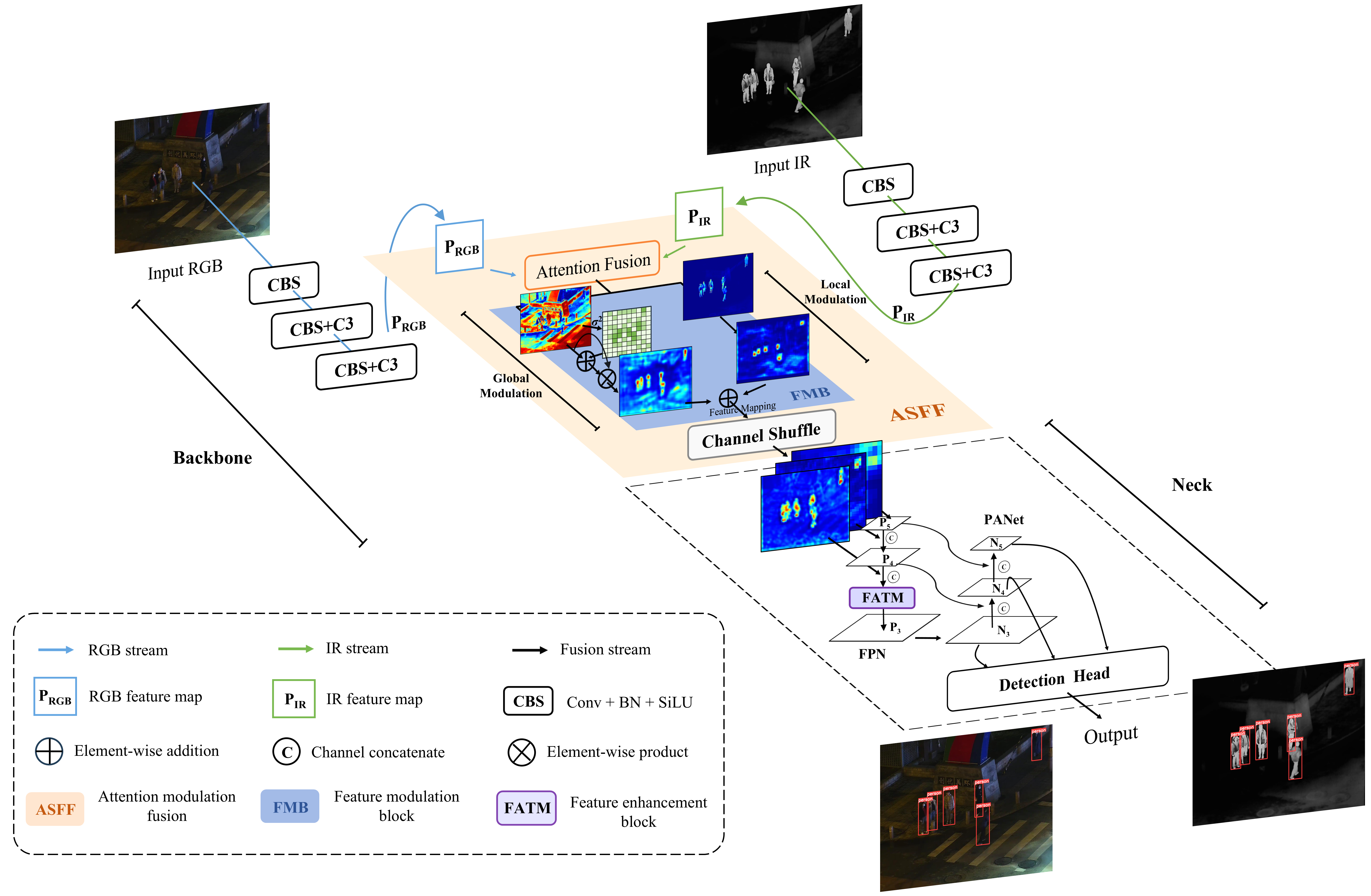}
\caption{The overall architecture of LASFNet. The detection network consists of three main components: the feature extraction backbone, the neck, and the detection head. In the feature extraction backbone, the single-feature-level fusion unit, ASFF, is responsible for fusing the \(\text{P}_{\text{RGB}}\) and \(\text{P}_{\text{IR}}\) features extracted by the dual-stream structure. In the Neck, we retain the feature pyramid network (FPN)\cite{lin2017feature} with a top-down structure and the path aggregation network (PANet)\cite{liu2018path} with a bottom-up structure, while introducing a feature attention transformation module (FATM) to enhance the network’s ability to focus on critical features, thereby reducing feature loss. }
\label{fig2}
 
\end{figure*}
  
\section{Related Work}\label{sec:related_work}
\noindent\emph{A. Multimodal Object Detection}

With significant advancements in object detection technology, including both one-stage\cite{Zheng2022EN} and two-stage detectors\cite{Wu2019fa}, visible-light-based object detection has made substantial progress as a key task in computer vision\cite{Cheng_2024_CVPR}. However, visible light data is susceptible to environmental factors that degrade its quality, such as adverse weather conditions like cloud cover, fog, and low-light environments, resulting in suboptimal visual performance\cite{Hnewa2021},\cite{Isaac-Medina_2021_ICCV}. To mitigate the limitations of single-modality detection, researchers have combined infrared data—highly complementary to visible light—into multimodal fusion approaches to enhance detection performance. Zhang et al.\cite{Zhang2023} employed super-resolution-assisted pixel-level fusion to achieve higher spatial resolution for remote sensing small object detection. However, this approach requires significant computational resources, leading to longer processing times. Wang et al.\cite{wang2024rgb} utilized decision-level fusion through multiple branches to mitigate the influence of individual modalities. However, the extensive use of branch structures not only increases computational overhead but may also introduce an imbalance in decision quality.  In contrast, feature-level fusion enables the flexible and in-depth extraction of rich feature representations, which leads to improved model performance. As a result, it has become a widely adopted approach. Zhang et al.\cite{zhang2021weakly} introduced an adjacent region similarity alignment strategy to resolve positional misalignment between modalities. Fu et al.\cite{fu2023lraf} proposed a feature-enhanced long-range attention fusion network to improve long-range dependencies across modalities, thus enhancing multimodal detection performance. However, these methods often do not deeply explore the correlations and complementarities between different modality features for data fusion. To address this, Zhang et al.\cite{zhang2023differential} proposed a differential feature perception network leveraging adversarial learning between infrared and visible light modalities to fully exploit complementary information. Additionally, Wang et al.\cite{wang2024cdc} designed a cross-modal enhancement strategy to capture modality correlations and introduced a dynamic convolution-based, cross-scale enhancement fusion guided by cross-modal loss to adaptively focus on bimodal features, achieving comprehensive feature generation. It is worth noting that while most approaches in the field achieve high detection accuracy through sophisticated  multimodal fusion, they often overlook the substantial computational cost associated with these fusion methods, which hinders the broader application of multimodal object detection. This paper presents the LASFNet lightweight network, which addresses the limitations of existing pixel-level fusion models that rely on auxiliary branches to extract spatial details but struggle to capture key features. We propose an innovative single-feature fusion strategy, utilizing a self-modulation mechanism to adaptively extract core information from the fused features. This approach preserves spatial detail perception while achieving simultaneous improvements in accuracy and speed at minimal computational cost, thereby optimizing the balance between efficiency and precision.

\vspace{1.5em}

\noindent\emph{B. Self-modulation in Multimodal Detection}

The core idea of self-modulation mechanisms\cite{zheng2025smfanet} is to enable networks to automatically adjust their internal activations or weights based on input features, thereby better capturing the structure and characteristics of the input data. The Transformer\cite{dosovitskiy2020image} captures dependencies between different positions in the input sequence through its self-attention mechanism, and dynamically associates information from different modalities, making it particularly well-suited for multi-modal information fusion\cite{dong2024seadate} to assist in multi-modal object detection. Yuan et al.\cite{Yuan2024Improving} proposed a complementary fusion Transformer that utilizes self-attention to address issues related to misalignment and inaccurate fusion. Helvig et al.\cite{helvig2024caff} combined cross-attention with the Transformer-based DINO detector to extract meaningful correlations between modalities. Furthermore, Vision Mamba\cite{zhu2024vision},\cite{Huang2024} (Vim) achieves the same global visual context modeling capability as the Transformer with lower computational complexity, showing significant potential for multi-modal object detection\cite{liu2024cross}. Li et al.\cite{li2024cfmw} introduced a cross-modal fusion Mamba for multi-spectral object detection under adverse weather conditions. Ren et al.\cite{ren2024remotedet} utilized the global perception advantage of Mamba in fusion tasks, combining it with local information to improve the separability of small remote-sensing objects. Given the significant computational overhead introduced by both Transformer and Mamba in enhancing multi-modal object detection, some researchers have cleverly combined attention information for self-modulation. Chen et al.\cite{Chen_2024_CVPR} employed attention and deformable convolutions to adaptively adjust multi-modal feature discrepancies, addressing weak misalignment issues in multi-modal detection. Wan et al.\cite{Wan2023} adaptively smoothed node representations using attention information to alleviate the object occlusion problem. Although the methods above have proven effective in multi-modal object detection tasks, their complexity remains substantial, hindering lightweight requirements. To address this, we propose a lightweight, attention-guided self-modulation feature fusion (ASFF) mechanism that efficiently integrates global and local multimodal information to enhance multimodal detection performance, with a lightweight design. 

\section{Methodology}\label{sec:method}
\noindent\emph{A. Architecture Overview}
\vspace{0.5em}

Figure \ref{fig2} illustrates the workflow of our proposed LASFNet for multimodal object detection, which consists of three main components: the backbone, the neck, and the detection head. Notably, we introduce two novel modules to enhance detection performance, enabling the network to achieve efficient, high-performance detection in a lightweight design. First, in contrast to previous multimodal fusion methods\cite{wang2024cross},\cite{althoupety2024daff},\cite{cao2023multimodal}, that employ multiple parallel feature-level fusion modules in the backbone, we use a single feature-level fusion model, ASFF, to progressively and adaptively fuse paired RGB-IR modality features, yielding a comprehensive and rich fused feature representation. Additionally, while maintaining robust multi-scale feature extraction through a top-down feature pyramid network (FPN) \cite{lin2017feature}and bottom-up path aggregation network (PANet)\cite{liu2018path} in the neck, we introduce a simple yet flexible feature attention transformation module (FATM) to improve feature focus and reduce information loss. Finally, the multi-scale features from the neck are passed to the detection head to obtain the final detection results. 

\vspace{1.5em}
\noindent\emph{B. Attention-guided Self-modulation Feature Fusion}
\vspace{0.5em}

\begin{figure*}[ht]
 
\centering
\includegraphics[width=\textwidth,keepaspectratio]{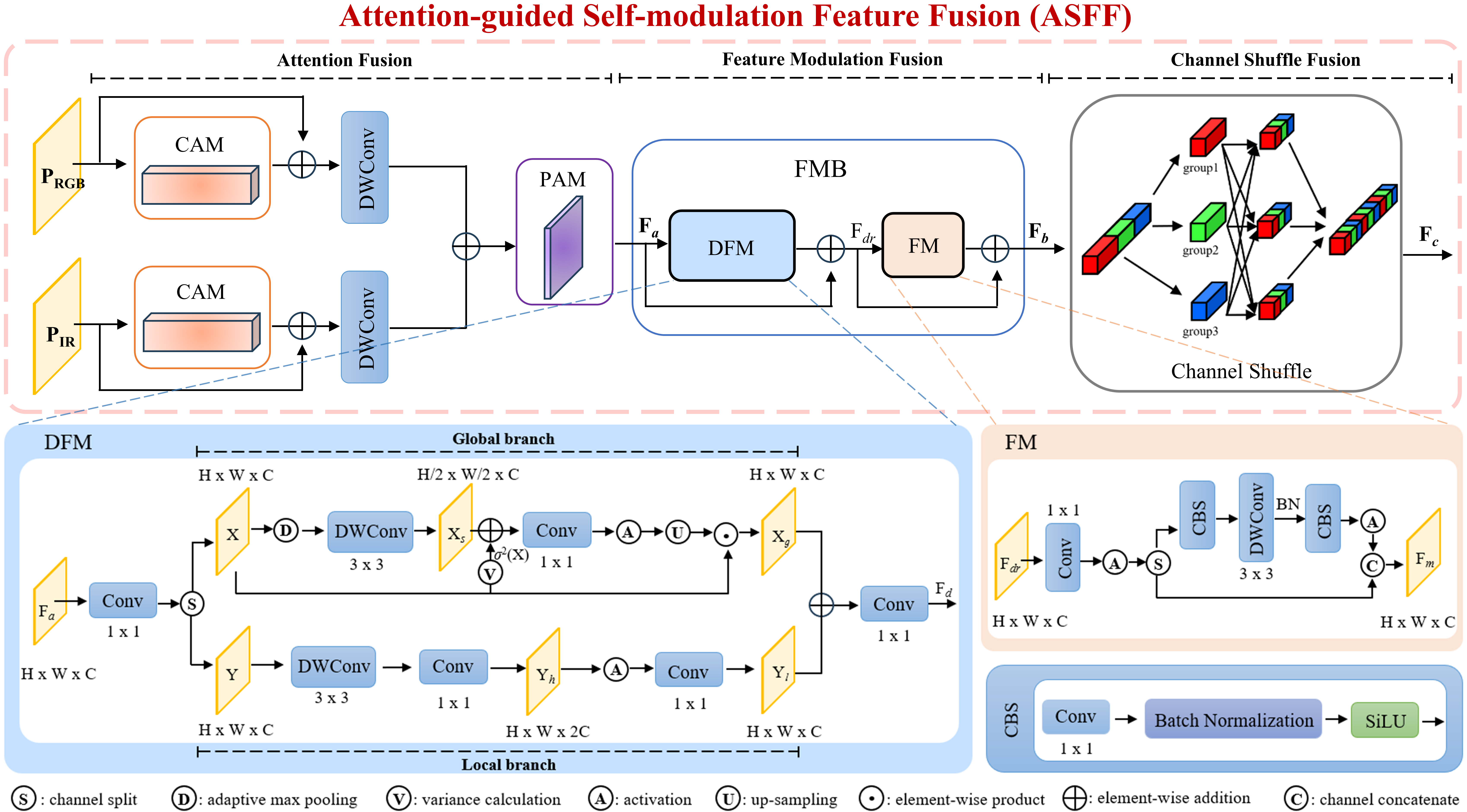}
\caption{Detailed architecture of our proposed attention-guided self-modulation feature fusion (ASFF) module. The fusion process is divided into three stages: attention fusion, feature modulation fusion, and channel shuffling fusion. The RGB and IR feature maps are the inputs of the whole fusion process, and the outputs of the three stages are \(\mathbf{F}_a\), \(\mathbf{F}_b\), and \(\mathbf{F}_c\).
 }
\label{fig3}
 
\end{figure*}

We propose the attention-guided self-modulation feature fusion (ASFF) module for comprehensive multimodal fusion through adaptive progressive feature integration. Figure \ref{fig3} illustrates the detailed structure of the proposed ASFF module, which consists of three fusion steps: attention fusion, feature modulation fusion, and channel shuffle fusion.

\textbf{Attention Fusion}: In multimodal fusion, attention mechanisms dynamically weight modality regions to focus on key information\mbox{\cite{zhou2024dpnet}.} In the proposed ASFF module, we first apply a dual-branch structure to the incoming RGB and IR feature maps \(\mathbf{P}_{RGB},\mathbf{P}_{IR}\in\mathbb{R}^{H\times W\times C}\) using a channel attention mechanism (\(\mathrm{CAM}\)). This mechanism adaptively weights feature channels by importance, generating enhanced multimodal representations\mbox{ \(\mathbf{M}_{RGB},\mathbf{M}_{IR}\in\mathbb{R}^{H\times W\times C}\).} The process is defined by: 
\begin{equation}
\begin{aligned}
    \mathbf{M}_{IR} &= \mathrm{CAM}(\mathbf{P}_{IR}), \\
    \mathbf{M}_{RGB} &= \mathrm{CAM}(\mathbf{P}_{RGB}), \\
    \mathrm{CAM}(x) &= x \odot \sigma \left( \text{Conv1d}\left( \mathrm{AP}(x) \right) \right),
\end{aligned}
\end{equation}
where \( \odot \) denotes element-wise multiplication, \(\mathrm{\sigma}(\cdot)\) represents the Sigmoid function, \(\mathrm{AP}(\cdot)\) and \(\text{Conv1d}(\cdot)\) correspond to Adaptive Average Pooling and 1D convolution operations, respectively. 
 
Subsequently, the channel-attended feature maps \(\mathrm{M}_{RGB}\) and \(\mathrm{M}_{IR}\) are element-wise added to their corresponding original inputs \(\mathrm{P}_{RGB}\) and \(\mathrm{P}_{IR}\) , forming residual connections. This operation retains the original spatial and channel information while incorporating the attention-adjusted weights, enhancing the model's focus on important features. The resulting sums are then passed through depthwise convolutions to preserve the independence of the channel features, yielding the enhanced modality feature maps  \(\hat{\mathbf{M}}_{{RGB}},\hat{\mathbf{M}}_{{IR}}\in\mathbb{R}^{H\times W\times C}\) , as defined by: 
\begin{equation}
\begin{split}
  \mathbf{\hat{M}}_{\mathit{IR}} &= \text{DWConv}(\mathbf{P}_{\mathit{IR}} + \mathbf{M}_{\mathit{IR}}), \\
  \mathbf{\hat{M}}_{\mathit{RGB}} &= \text{DWConv}(\mathbf{P}_{\mathit{RGB}} + \mathbf{M}_{\mathit{RGB}}),
\end{split}
\end{equation}
where \(\text{DWConv}\) denotes depthwise convolutions with a 3×3 kernel. We fuse the enhanced RGB and IR feature maps (\mbox{\(\hat{\mathbf{M}}_{{RGB}},\hat{\mathbf{M}}_{{IR}}\)}) to improve cross-modal complementarity via element-wise addition. A positional attention mechanism (PAM) then spatially weights features along horizontal/vertical axes, amplifying discriminative local features. The final attention-fused feature map \(\mathbf{F}_a\) is computed as: 
\begin{equation}
\begin{aligned}
    \mathbf{F}_a &=\mathrm{PAM}(\hat{\mathbf{M}}_{{RGB}} +\hat{\mathbf{M}}_{{IR}}),\\
\mathrm{PAM}(x) &=x\odot\mathrm{Att}_h(x)\odot\mathrm{Att}_v(x),\\
\mathrm{Att}_i(x) &=\sigma(\text{Conv}(\mathrm{Avg}_i(x))),
\end{aligned}
\end{equation}
where \(\mathrm{Att}_h(\cdot)\) computes the horizontal position attention weights by applying average pooling \(\mathrm{Avg}_h(\cdot)\) along the width of the input feature map. \(\mathrm{Att}_v(\cdot)\) computes the vertical position attention weights using average pooling \(\mathrm{Avg}_v(\cdot)\) along the height of the input feature map. 

\textbf{Feature Modulation Fusion}: The feature modulation block (FMB) synergizes dynamic feature modulation (DFM) and feature mapping (FM). This architecture strengthens multi-scale representations (global/local) and optimizes feature learning in attention-fused patterns. Specifically, DFM adapts the input features through dynamic modulation to better capture the relationships between global and local features. The process begins by normalizing the attention-fused features \(\mathbf{F}_a\), followed by a 1×1 convolution to expand the channels. The channels are then split into two branches for further processing. This operation can be mathematically expressed as: 
\begin{equation}
    \{\mathbf{X},\mathbf{Y}\}=S(\text{Conv}(\|\mathbf{F}_a\|_2)),
\end{equation}
where \(\|\cdot\|_2\) denotes L2 normalization, Conv represents a 1×1 convolutional layer, and \(S(\cdot)\) refers to the channel splitting operation. Then, the global and local branches process \(\text{X and Y}\in\mathbb{R}^{H\times W\times C}\) in parallel to generate global features \(\mathbf{X}_{g}\) and local features \(\mathbf{Y}_{l}\). 

In the global branch, we first apply adaptive max pooling to downsample the input and obtain salient representations, then feed it into a 3×3 \(\text{DWConv}\) to capture the global structural information \(\mathbf{X}_s\in\mathbb{R}^{H/2\times W/2\times C}\):
\begin{equation}
    \mathbf{X}_s=\text{DWConv}(D(\mathbf{X})),
\end{equation}
where \(D(\cdot)\) denotes the adaptive max pooling operation with a scale factor of 2. To embed the global feature information and adaptively adjust the global structural information \(\mathbf{X}_{s}\), we introduce the variance \(\sigma^2(\mathbf{X})\) to provide spatial variability for feature fusion, helping the model focus on feature variations in different regions. Two learnable parameters \(\alpha \) and \(\beta \) allow the model to dynamically adjust the contributions of the variance and \(\mathbf{X}_{s}\), enabling adaptive fusion of information from different spatial regions. The result is then merged using a 1×1 convolution: 
\begin{equation}
\renewcommand{\arraystretch}{1.5}  % 设置行间距
\begin{split}
    \sigma^2(\mathbf{X}) &= \frac{1}{N}\sum_{i=0}^{N-1}(x_i-\mu)^2, \\
    \mathbf{X}_m &= \text{Conv}(\mathbf{X}_s\cdot\alpha + \sigma^2(\mathbf{X})\cdot\beta),
\end{split}
\end{equation}
where \(\sigma^2(\mathbf{X})\in\mathbb{R}^{1\times1\times C}\) is the variance of  \(\mathbf{X}\).  \(\mathit{N}\) is the total number of pixels, \(x_{i}\) represents each pixel value, and \(\mu \) is the mean of all pixel values. \(\mathbf{X}_m\in\mathbb{R}^{H\times W\times C}\) represents the globally-modulated feature. Finally, the modulated feature \(\mathbf{X}_{m}\) is used to adjust the original input \(\mathbf{X}\) globally, producing the representative global branch feature \(\mathbf{X}_{g}\in\mathbb{R}^{H\times W\times C}\):
\begin{equation}
    \mathbf{X}_g=\mathbf{X}\odot U(\phi(\mathbf{X}_m)),
\end{equation}
where \(\phi(\cdot)\) represents the GELU activation function, \(U(\cdot)\) denotes the nearest neighbor upsampling operation. 

Local detail features complement the global features to provide more contextual information and enhance the model's expressiveness. We design a simple bottleneck-inspired local branch to capture local features. Specifically, we first perform a dimensionality increase using a 3×3 depthwise convolution to encode local information in \(\text{Y}\), resulting in \(\mathbf{Y}_{h}\). Then, we apply 1×1 convolutions with GELU activations to reduce the dimensionality and enhance the local features, \(\mathbf{Y}_{l}\in\mathbb{R}^{H\times W\times C}\), as follows: 
\begin{align}
\mathbf{Y}_h &= \text{Conv}(\text{DWConv}(\mathbf{Y})), \\
\mathbf{Y}_l &= \text{Conv}(\phi(\mathbf{Y}_h)),
\end{align}
where \(\mathbf{Y}_h\in\mathbb{R}^{H\times W\times2C}\) represents the locally encoded information after dimensionality expansion. Finally, the global and local features are fused through element-wise addition and passed through a 1×1 convolution to produce the representative output of the DFM module \(\mathbf{F}_d\in\mathbb{R}^{H\times W\times C}\), as follows:
\begin{equation}
    \mathbf{F}_d=\text{Conv}(\mathbf{X}_g+\mathbf{Y}_l),
\end{equation}
where  \(\mathbf{X}_{g}\) and \(\mathbf{Y}_{l}\) are the global and local features, respectively. 

The feature mapping (FM) module further refines the output features of the DFM module through different channel-wise processing, enhancing the ability to capture both global and local features while avoiding excessive computational overhead. As shown in Figure \ref{fig3}, the FM module first employs a 1×1 convolution with GELU activation to perform dimensionality expansion, thereby increasing the feature channels to facilitate information interaction. The expanded features are then split into two parts, \(\{\mathbf{F}_{1},\mathbf{F}_{2}\}\), the features with fewer channels, \(\mathbf{F}_1\), undergo local encoding, while \(\mathbf{F}_2\) remains unchanged to preserve more global information. The processed  \(\mathbf{F}_1\) and  \(\mathbf{F}_2\) are concatenated and passed through a final 1×1 convolution to fuse the features and obtain the representative output \(\mathbf{F}_{m}\), restoring the dimensions to the original input state. This process is described by: 
\begin{equation}
\renewcommand{\arraystretch}{1.5}  % 设置行间距
\begin{split}
    \{\mathbf{F}_1,\mathbf{F}_2\}&=S(\phi(\text{Conv}(\|\mathbf{F}_{dr}\|_2))),\\
    \quad{W(x)}&=(\text{CBS}(\text{BN}(\text{DWConv}(\text{CBS}(x))))),\\
    \quad\mathbf{F}_m &=\text{Conv}(\text{Concat}[\phi({W}(\mathbf{F}_1)),\mathbf{F}_2])),
\end{split}
\end{equation}
where \(\mathbf{F}_{1}\in\mathbb{R}^{H\times W\times C/2}\), \(\mathbf{F}_2\in\mathbb{R}^{H\times W\times3C/2}\), \(W(\cdot)\) denotes the local encoding operation applied to \(\mathbf{F}_1\) and \(\mathrm{Concat}(\cdot)\) denotes channel concatenation. \(\text{CBS}(\cdot)\) refers to a sequence of 1×1 convolution, normalization, and SiLU activation, while BN denotes the batch normalization operation. As described earlier, the DFM and FM are combined into a feature modulation block (FMB) for a second feature fusion, generating a rich representation that combines spatial, channel, and both global and local multi-scale features. To stabilize model training and retain more informative features, residual connections are incorporated in the FMB. The final output \(\mathbf{F}_{b}\) of the FMB module is computed as: 
\begin{equation}
\begin{aligned}
\mathbf{F}_{dr} &= \mathrm{DFM}(\mathbf{F}_a) + \mathbf{F}_a, \\
\mathbf{F}_b  &= \mathrm{FM}(\mathbf{F}_{dr}) + \mathbf{F}_{dr}.
\end{aligned}
\end{equation}

\textbf{\textbf{Channel Shuffle Fusion} }: Channel shuffle\cite{zhang2018shufflenet},\cite{ma2018shufflenet} can be viewed as a lightweight channel reorganization operation. In traditional convolutional neural networks, convolution operations extract local features through sliding windows in the image space; however, interactions between channels are typically limited. The Channel shuffle operation breaks the independence between channels by shuffling the order of channels across different groups, enabling the network to exchange information between multiple channel groups. This approach promotes the complementarity between different features at a low computational cost, thus improving the model's representational power. The features obtained after channel shuffling, \(\mathbf{F}_{c}\), can be computed as follows: 
\begin{equation}
\begin{aligned}
    \mathbf{x}^{\prime} &= \text{reshape}(\mathbf{x}, B, G, C/G, H, W), \\
    \mathbf{x}^{\prime\prime} &= \text{transpose}(\mathbf{x}^{\prime}, 1, 2), \\
    \mathbf{F}_{c} &= \text{reshape}(\mathbf{x}^{\prime\prime}, B, C, H, W),
\end{aligned}
\end{equation}
where the reshape operation refers to a feature reshaping process, \(\mathbf{x}\in\mathbb{R}^{B\times C\times H\times W}\), with \({B}\) representing the batch size, \({G}\) being the number of channel groups, and \({C/G}\) denoting the number of channels per group. The reshaped tensor \(\mathbf{x}^{\prime}\) has the shape \((B,G,C/G,H,W)\) which can be interpreted as dividing \({C}\) channels into \({G}\) groups, each containing \({C/G}\) channels. The transpose operation swaps the group dimension \({G}\) with the channel dimension \({C/G}\) , thus enabling the elements of each channel group to be distributed across different groups, increasing the interaction between channels. After reshaping back to the original format, the final output of the channel shuffle fusion is obtained as \(\mathbf{F}_{c}\). Algorithm \mbox{\ref{alg:ASFF}} shows the process involved in the ASFF module.

ASFF reduces computational overhead with an efficient feature fusion strategy, using attention weighting and dynamic modulation to retain key features, thereby enhancing multimodal feature learning. This lightweight design offers significant advantages in multimodal object detection, especially in embedded or resource-constrained environments, balancing computational efficiency and performance. 

\begin{algorithm}[htbp]
\caption{ASFF Module} \label{alg:ASFF}
\SetAlgoLined
\KwIn{Pairwise visible/infrared features: $\mathbf{P}_{RGB}$, $\mathbf{P}_{IR}$}
\KwOut{Fusion feature $\mathbf{F}_c$}
 
\textbf{Stage 1: Attention Fusion}\\
1. Obtain channel-weighted features via channel attention:\\
$\mathbf{M}_{RGB}, \mathbf{M}_{IR} \leftarrow \mathrm{CAM}(\mathbf{P}_{RGB}), \mathrm{CAM}(\mathbf{P}_{IR})$ \\
2. Enhance features via residual connection: \\
$\hat{\mathbf{M}}_{RGB} \leftarrow \text{DWConv}(\mathbf{M}_{RGB} + \mathbf{P}_{RGB})$ \\
$\hat{\mathbf{M}}_{IR} \leftarrow \text{DWConv}(\mathbf{P}_{\mathit{IR}} + \mathbf{M}_{\mathit{IR}})$ \\
3. Generate attention fusion feature: \\
$\mathbf{F}_a \leftarrow \mathrm{PAM}(\hat{\mathbf{M}}_{{RGB}} +\hat{\mathbf{M}}_{{IR}})$ \\
\textbf{Stage 2: Feature Modulation Fusion}\\
4. Split $\mathbf{F}_a$ into two branches: \\
\hspace*{5mm} Global branch: $\mathbf{X}_g \leftarrow \text{GlobalProcess}(\mathbf{X})$ \\
\hspace*{5mm} Local branch: $\mathbf{Y}_l \leftarrow \text{LocalProcess}(\mathbf{Y})$ \\
5. Fuse two branches: \\
$\mathbf{F}_d \leftarrow \text{Conv}(\mathbf{X}_g+\mathbf{Y}_l)$ \\
6. Obtain modulated feature via residual: \\
$\mathbf{F}_{dr} \leftarrow \mathbf{F}_d + \mathbf{F}_a$ \\
$\mathbf{F}_b \leftarrow \mathrm{FM}(\mathbf{F}_{dr}) + \mathbf{F}_{dr}$ \\
\textbf{Stage 3: Channel Shuffle Fusion}\\
7. Reshape feature: $\mathbf{x}' \leftarrow \text{reshape}(\mathbf{F}_b)$ \\
8. Transpose for interaction: $\mathbf{x}'' \leftarrow \text{transpose}(\mathbf{x}')$ \\
9. Recover shape: $\mathbf{F}_c \leftarrow \text{reshape}(\mathbf{x}'')$ \\
\Return $\mathbf{F}_c$
\end{algorithm}

\begin{figure}[ht]
 
\centering
\includegraphics[width=\linewidth]{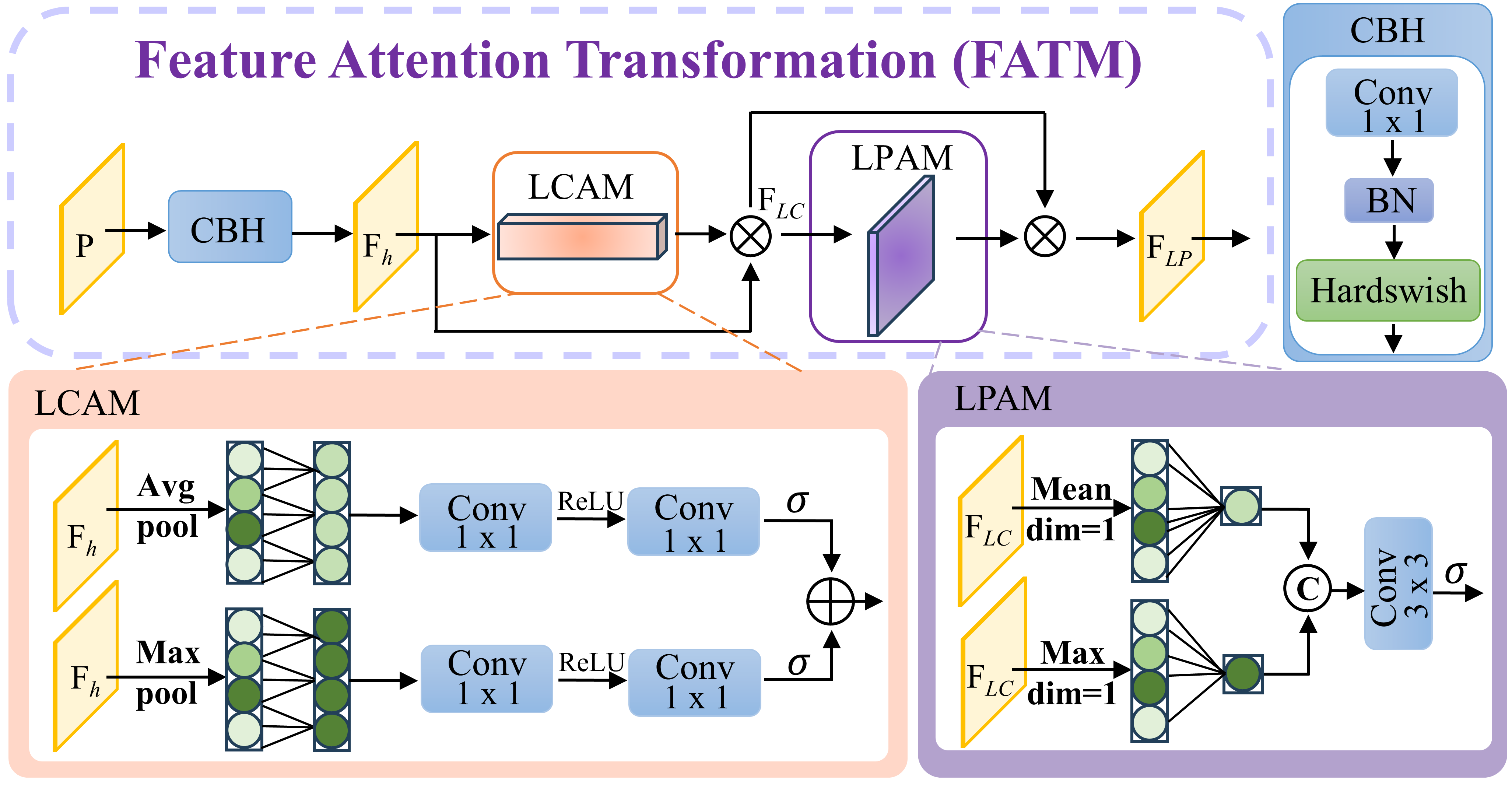}
\caption{Detailed architecture of our proposed Feature Attention Transformation Module (FATM). The lightweight channel attention module (LCAM) and the lightweight positional attention module (LPAM), synergistically preserve critical feature representations while effectively mitigating information degradation throughout the process.}
\label{fig4}
 
\end{figure}

\vspace{1.5em}
\noindent\emph{C. Feature Attention Transformation Module}
\vspace{0.5em}

Multi-stage sampling and fusion in the neck may blur backbone-extracted features while introducing redundancy and noise, degrading detection quality. To address this, we propose a lightweight feature attention transformation module (FATM) at the FPN's end. By applying channel and positional weighting, FATM suppresses irrelevant information while preserving critical features. This enhancement helps subsequent PAN networks better integrate multi-scale features, ultimately improving detection performance.

The FATM first processes input through a conv-bn-hardswish (CBH) block, leveraging depthwise convolution to boost nonlinear modeling capacity while maintaining computational efficiency, as shown in Figure \mbox{\ref{fig4}.} The operation can be mathematically represented as follows: 
\begin{equation}
    \mathbf{F}_{h}=\text{Hardswish}(\text{BN}(\text{Conv}3\text{×}3(\mathbf{P}))),
\end{equation}
where \(\text{Hardswish}(\cdot)\) represents the Hardswish nonlinear activation function, and \(\mathbf{P}\) denotes the input features. Subsequently, the extracted features \(\mathbf{F}_{h}\) are fed into the lightweight channel attention module (LCAM) to adjust the importance of different channels. This enables the model to dynamically adjust the weight of each channel based on its contribution within the input feature map, thus strengthening the network's attention to useful channel information. The channel-attention-enhanced feature \(\mathbf{F}_{LC}\) is given by:
\vspace{0.25em}
\begin{equation}
   \mathbf{F}_{LC}=\mathbf{F}_{h}\odot\mathrm{LCAM}(\mathbf{F}_{h}),
\end{equation}
\vspace{0.25em}where the \(\mathrm{LCAM}(\cdot)\) extracts attention information from the global average and local maximum responses of the input features using a dual-branch structure. The final channel attention weights are obtained by fusing this information, and the operation is expressed as: 
\vspace{0.25em}
\begin{equation}
\renewcommand{\arraystretch}{1.5}  % 设置行间距
\begin{aligned}
    \mathrm{LCAM}(\mathbf{F}_{h}) &= \sigma\left( \text{Conv} \left( \mathrm{ReLU} \left( \text{Conv} \left( \mathrm{AP}(\mathbf{F}_{h}) \right) \right) \right) \right) \\
    &\hspace{-1em} + \sigma\left( \text{Conv} \left( \mathrm{ReLU} \left( \text{Conv} \left( \mathrm{MP}(\mathbf{F}_{h}) \right) \right) \right) \right),
\end{aligned}
\end{equation}
\vspace{0.25em}where \(\mathrm{\sigma}(\cdot)\) denotes the Sigmoid function, and \(\mathrm{ReLU}(\cdot)\) represents the ReLU activation function. \(\mathrm{AP}(\cdot)\) and \(\mathrm{MP}(\cdot)\) refer to Adaptive Average Pooling and Adaptive Max Pooling, respectively. After applying channel attention, the weighted features are passed into the lightweight positional attention module (LPAM), which adaptively focuses on important spatial regions, thus enabling dynamic adjustment of the positional information in the feature map. The positional attention-enhanced features \(\mathbf{F}_{LP}\) is represented as the final output as follows: 
\vspace{0.25em}
\begin{equation}
\renewcommand{\arraystretch}{1.5}  % 设置行间距
    \mathbf{F}_{LP}=\mathbf{F}_{LC}\odot\mathrm{LPAM}(\mathbf{F}_{LC}),
\end{equation}
\vspace{0.25em}where \(\mathrm{LPAM}(\cdot)\) performs global average pooling and global max pooling along the channel dimension in a dual-branch manner, followed by fusion to obtain the final position attention weights, which can be expressed as: 
\vspace{0.25em}
\begin{align}
\mathrm{LPAM}(\mathbf{F}_{LC}) &= \sigma\left(\text{Conv}_{3 \times 3}\left(\text{Concat}\left[\mathbf{F}_{{Max}}, \mathbf{F}_{{Mean}}\right]\right)\right)\!
\end{align}
\vspace{0.25em}where \(\mathbf{F}_{Max}\) denotes the result of applying global max pooling along the channel dimension of \(\mathbf{F}_{LC}\), while \(\mathbf{F}_{Mean}\) denotes the result of applying global average pooling along the channel dimension of \(\mathbf{F}_{LC}\). Algorithm\mbox{ \ref{alg:FATM}} shows the process involved in the FATM module.

The FATM module successfully achieves a dual goal of lightweight design and feature enhancement by cleverly combining channel and positional attention mechanisms. It significantly improves the model's ability to selectively focus on key features while maintaining low computational overhead and a fewer number of parameters.

\begin{algorithm}[htbp]
\caption{FATM Module} \label{alg:FATM}
\SetAlgoLined
\KwIn{Feature map $\mathbf{P}$}
\KwOut{Enhanced feature $\mathbf{F}_{LP}$}
1. Extract features via CBH: \\
$\mathbf{F}_h \leftarrow \text{CBH}(\mathbf{P})$ \\
2. Dual-branch channel attention LCAM: \\
\hspace*{5mm} AAP branch: $\mathbf{W}_g \leftarrow \sigma(\text{CRC}(\text{AP}(\mathbf{F}_h)))$ \\
\hspace*{5mm} AMP branch: $\mathbf{W}_l \leftarrow \sigma(\text{CRC}(\text{MP}(\mathbf{F}_h)))$ \\
3. Fuse channel weights: \\
$\mathbf{F}_{LC} \leftarrow \mathbf{F}_{h}\odot(\mathbf{W}_g + \mathbf{W}_l)$ \\
4. Dual-branch position attention LPAM: \\
\hspace*{5mm} GAP branch: $\mathbf{S}_g \leftarrow \text{GAP}(\mathbf{F}_{LC})$ \\
\hspace*{5mm} GMP branch: $\mathbf{S}_m \leftarrow \text{GMP}(\mathbf{F}_{LC})$ \\
5. Fuse position weights: \\
$\mathbf{F}_{LP} \leftarrow  \mathbf{F}_{LC}\odot\sigma\left(\text{Conv}\left(\text{Concat}\left[\mathbf{S}_g ,\mathbf{S}_m\right]\right)\right)\!$ \\
\Return $\mathbf{F}_{LP}$
\end{algorithm}

\vspace{0.5em}
\noindent\emph{D. Loss Function}
\vspace{0.5em}

The overall loss of the model remains consistent with the total loss used in YOLOv5\cite{jocher2022ultralytics}, consisting of three components: the bounding box regression loss, object classification loss, and object confidence loss. The final total loss can be expressed as: 
\vspace{0.25em}
\begin{equation}
\renewcommand{\arraystretch}{1.5}  % 设置行间距
    L_\mathrm{total}=\lambda_\mathrm{box}\cdot L_\mathrm{box}+\lambda_\mathrm{obj}\cdot L_\mathrm{obj}+\lambda_\mathrm{cls}\cdot L_\mathrm{cls},
\end{equation}
\vspace{0.25em}where the hyperparameters \(\lambda_{\mathrm{box}}\), \(\lambda_{\mathrm{obj}}\), \(\lambda_{\mathrm{cls}}\) are used to adjust the relative importance of each loss component.

\section{Experiments}\label{sec:experiments}
\noindent\emph{A. Datasets}
\vspace{0.5em}

We conducted experiments on the DroneVehicle, LLVIP, and VTUAV\_det datasets to evaluate the performance of LASFNet. 

1) \textbf{DroneVehicle}\cite{sun2022drone}: This dataset is primarily designed for visible-infrared (RGB-IR) modality-based aerial vehicle detection, and includes images captured at various times throughout the day. The objects in this dataset are relatively small, and challenges include poor object visibility in the visible modality during late-night hours, as well as thermal artifacts resembling vehicles in the infrared modality. The dataset consists of five object classes, with a total of 19,459 RGB-IR image pairs, of which 17,990 pairs are used for training and the remaining 1,469 pairs are used for testing. Due to the substantial differences between the visible and infrared modalities, this dataset is particularly valuable for evaluating the model's performance in complementary data fusion.

2) \textbf{LLVIP}\cite{jia2021llvip}: This dataset is a large-scale pedestrian detection dataset captured under low-light conditions using a binocular camera. It contains 15,488 pairs of strictly aligned visible-infrared images, with 12,025 pairs used for training and 3,463 pairs used for testing. The dataset has undergone rigorous alignment, and the object scales in the images are relatively consistent compared to the other two datasets, with a generally larger object scale. This makes it suitable for assessing the model's overall detection performance.

3) \textbf{VTUAV\_det}: This dataset is derived from the VTUAV\cite{zhang2022visible} dataset, which was originally created for large-scale multi-modal drone tracking, and has been re-annotated by Zhang et al.\cite{zhang2023drone} for multi-modal object detection. It includes 11,392 image pairs used for training, with 5,378 image pairs reserved for testing. The dataset contains objects of three different sizes (small, medium, and large). Moreover, the images are not strictly aligned, and the dataset contains challenges such as significant object occlusion and image blur. Therefore, this dataset is highly suitable for evaluating the model's performance in handling challenges related to multi-scale variations, object occlusion, and image alignment.
\vspace{1.5em}

\noindent\emph{B. Implementation Details}
\vspace{0.5em}

The proposed method introduces a novel feature extraction backbone network built upon CSPDarknet53 from YOLOv5 in PyTorch. After extracting feature maps from each branch in the dual-branch architecture, a single fusion operation is performed. The model is trained using the SGD optimizer on dual NVIDIA GeForce RTX 3090 GPUs. During training on the DroneVehicle dataset, we converted the original oriented bounding boxes to horizontal bounding boxes for experimentation. The network was trained with a batch size of 48, an initial learning rate of 0.01, momentum set to 0.937, and weight decay set to 0.0005, with a total of 300 epochs. When calculating inference speed, all methods were executed on a single GPU with a unified batch size of 1.

\begin{table}[]
    \centering
    \caption{Comparison of the Performance of Multi-Feature Fusion Baseline and the Proposed Single-Feature Fusion Baseline on the LLVIP Dataset.}
    \label{tab:my_label1}
    \begin{tabular}{ccccc} 
        \toprule
        Method & mAP50 & mAP & Params (M) & GFLOPs \\ 
        \midrule
        Multi Fusion Baseline& \textbf{0.965} & 0.643 & 11.2 & 26.1 \\
        \\[-0.2cm]
        Single Fusion Baseline& 0.964 & \textbf{0.644} & \textbf{7.2}  & \textbf{20.8} \\ 
        \bottomrule
    \end{tabular}
\end{table}
\vspace{1.0em}
\noindent\emph{C. Accuracy Metrics}
\vspace{0.5em}

We evaluate the proposed method using commonly adopted detection metrics in the field of object detection, including mean Average Precision (mAP) and mAP50, along with giga floating point operations per second (GFLOPs) and parameter size (Params) for model efficiency assessment. The mAP is a key metric for evaluating the detection performance of a model. It measures the precision of an object detection model across multiple classes and different intersection over union (IoU) thresholds, providing a comprehensive and rigorous evaluation standard. Specifically, mAP50 refers to the mean average precision computed at an IoU threshold of 0.5, which is typically used as a baseline evaluation metric for detection tasks. GFLOPs is a crucial indicator for assessing the computational demand and efficiency of a model during inference. Params refers to the storage requirement of the model, serving as an essential metric for evaluating the model's feasibility and deployment challenges on storage-constrained devices.

\vspace{1.5em}
\noindent\emph{D. Ablation Experiments}
\vspace{0.5em}

In this section, we evaluate the effectiveness of the proposed single-feature fusion baseline network and perform an ablation study by configuring the proposed modules in different ways. Since the LLVIP dataset has undergone rigorous alignment and provides higher image resolution, it ensures the reliability of model performance evaluation across different configurations. Therefore, we chose the LLVIP dataset for the ablation study. First, we compare the detection performance of the popular multi-feature fusion baseline with that of the proposed single-feature fusion baseline to assess the feasibility of the single-feature fusion approach. The two baseline fusion methods are shown in Figure \ref{fig1}. Additionally, we investigate the performance improvements brought by the proposed ASFF and FATM modules within the baseline network. The following subsections provide a detailed analysis of these results. 

1) Effectiveness of the proposed baseline: We compare the detection results and model performance of two fusion baselines in Table \ref{tab:my_label1}, where both baseline methods use element-wise addition for fusion. In terms of detection accuracy, the performance of the single-feature fusion baseline is comparable to that of the multi-feature fusion baseline. However, in terms of model efficiency, the single-feature fusion baseline reduces the number of Params by approximately 36\% and reduces the computational cost (GFLOPs) by 20\% compared to the multi-feature fusion baseline. Therefore, the proposed single-feature fusion baseline demonstrates equivalent detection performance with a more efficient design, validating its effectiveness as a new baseline for multi-modal detection.  

2) Effectiveness of ASFF: In this experiment, we evaluate the performance of the three components of the ASFF module: attention fusion (Att), feature modulation block (FMB), and channel shuffle fusion (CS). Table \ref{tab:my_label2} presents a performance comparison of different component combinations, with the first row showing the baseline results without ASFF and the last row displaying the complete ASFF results. In the absence of attention fusion, we replace it with direct element-wise summation. As shown in the first four rows of Table \ref{tab:my_label2}, the application of each individual fusion component significantly improves detection performance, indicating that all three components contribute to multimodal information fusion. 

\begin{table}[]
    \centering
\caption{Performance Comparision Of ASFF Module With Different Components on the LLVIP Dataset.}
\label{tab:my_label2}
    \begin{tabular}{ccccccc}
        \toprule
         Att&  FMB&  CS&  mAP50&  mAP&  Params (M)& GFLOPs\\
         \midrule
         \ding{53}&  \ding{53}&  \ding{53}&  0.964&  0.644&  \textbf{7.2}& \textbf{20.8}\\
         \\[-0.2cm]
         \ding{51}&  \ding{53}&  \ding{53}&  0.965&  0.646&  \textbf{7.2}& \textbf{20.8}\\
         \\[-0.2cm]
         \ding{53}&  \ding{51}&  \ding{53}&  0.965&  0.648&  7.5& 23.7\\
         \\[-0.2cm]
         \ding{53}&  \ding{53}&  \ding{51}&  0.966&  0.646&  \textbf{7.2}& \textbf{20.8}\\
         \\[-0.2cm]
         \ding{51}&  \ding{51}&  \ding{53}&  0.965&  0.659&  7.5& 23.7\\
         \\[-0.2cm]
         \ding{53}&  \ding{51}&  \ding{51}&  0.967&  0.649&  7.5& 23.7\\
         \\[-0.2cm]
         \ding{51}&  \ding{53}&  \ding{51}&  0.967&  0.652&  \textbf{7.2}& \textbf{20.8}\\
         \\[-0.2cm]
         \ding{51}&  \ding{51}&  \ding{51}&  \textbf{0.969}&  \textbf{0.667}&  7.5& 23.7\\
         \bottomrule
    \end{tabular}

\end{table}
\begin{table}
    \centering
\caption{Performance Comparision Of LASFNet With Different Components on the LLVIP Dataset.}
\label{tab:my_label3}
    \begin{tabular}{cccccc}
        \toprule
         ASFF&  FATM&  mAP50&  mAP&  Params (M)& GFLOPs\\
         \midrule
         \ding{53}&  \ding{53}&  0.964&  0.652&  \textbf{7.2}& \textbf{20.8}\\
          \\[-0.2cm]
         \ding{53}&  \ding{51}&  0.965&  0.657&  7.5& 23.6\\
          \\[-0.2cm]
         \ding{51}&  \ding{53}&  0.969&  0.667&  7.5& 23.7\\
          \\[-0.2cm]
         \ding{51}&  \ding{51}&  \textbf{0.974}&  \textbf{0.676}&  7.7& 26.6\\
         \bottomrule
    \end{tabular}

\end{table}
The experimental results in the last four rows of Table \ref{tab:my_label2} show that pairwise combinations of Att, FMB, and CS lead to better performance than using each component individually. When all three components are combined into the complete ASFF module, the mAP improves by 2.3\% compared to the baseline. This demonstrates that the three fusion components complement each other, effectively enhancing the extraction of fused information and generating comprehensive and rich representative features. Attention fusion (Att) guides the model to focus on the most valuable information from different modalities in terms of both channels and spatial locations, providing strongly correlated input for the feature modulation block (FMB). After applying Att, FMB effectively adapts and adjusts the fused information at both global and local feature levels, facilitating multimodal fusion. Meanwhile, channel shuffle (CS) enhances the interaction of fused information without adding computational burden, allowing the network to extract meaningful features from multimodal data better. These results fully validate the lightweight and effective nature of the proposed ASFF for multimodal feature fusion. 

3) Effectiveness of FATM: This section examines the contribution of the feature attention transformation module (FATM) to model performance. We compare four experimental configurations: the first two rows of Table \ref{tab:my_label3} show the results with and without the FATM module in the proposed baseline, and the last two rows present the results when the FATM module is either removed or retained in the baseline framework with the ASFF module. In the absence of ASFF, element-wise summation was used as an alternative. From the comparison of the first two rows, it is clear that adding the FATM module to the baseline increases the overall mAP by 0.5\% with a minimal increase in parameters. Similarly, the analysis of the last two rows demonstrates that incorporating the FATM module into the ASFF-based fusion framework improves the overall mAP by nearly 1.0\%, while maintaining a lightweight architecture. These results confirm that the proposed FATM module effectively enhances the model's multiscale detection performance in a lightweight manner. 

\vspace{1.5em}
\noindent\emph{E. Comparison with State-of-the-Art Methods}
\vspace{0.5em}

\begin{figure*}[t!]
        \centering
        \begin{subfigure}
        \centering
    \includegraphics[width=\linewidth]{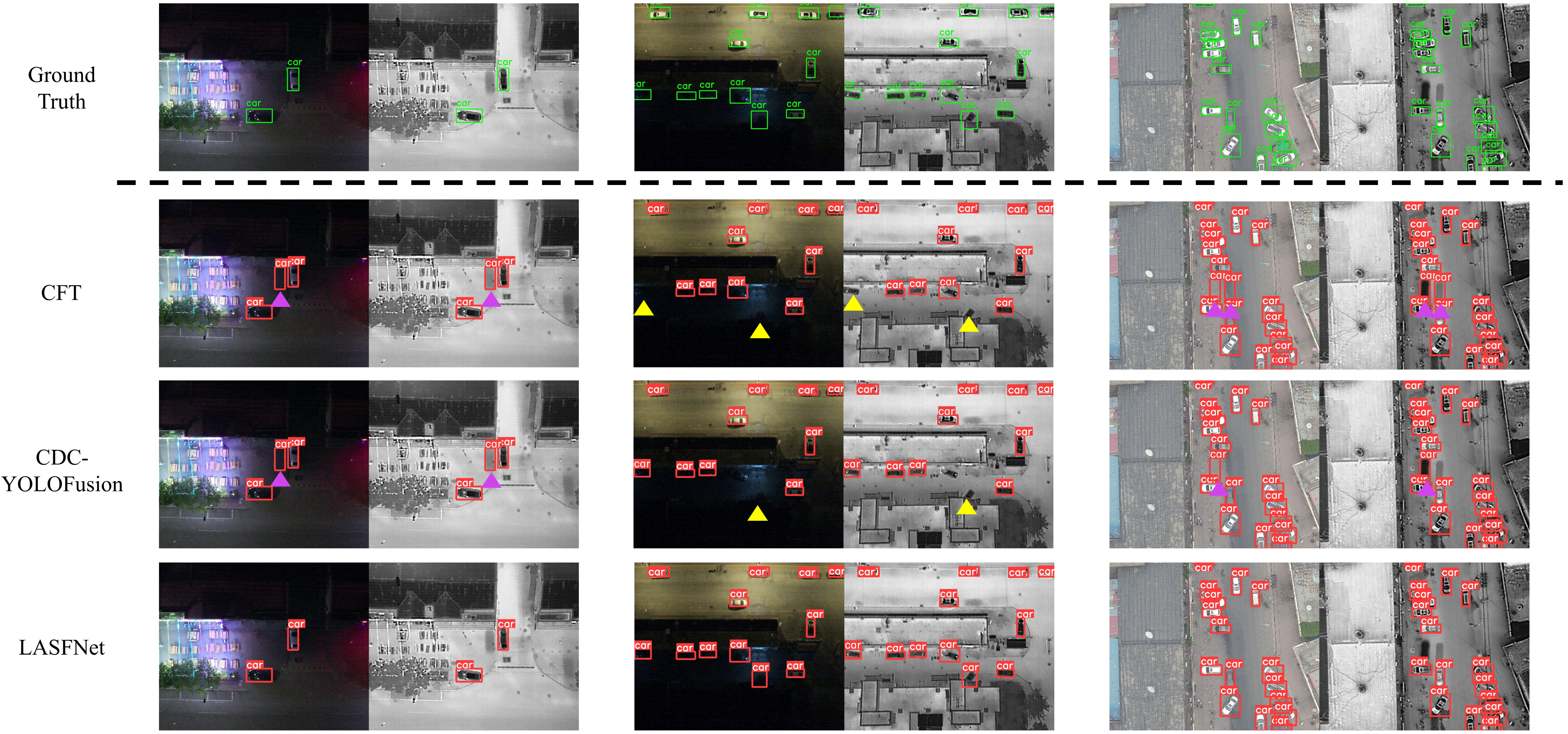}
    \caption{Three examples on DroneVehicle dataset to illustrate the detection results, where the purple triangles represent false positive objects, and the yellow triangles indicate the missing objects.}
    \label{fig6}
        \end{subfigure}
        \hfill
        \vspace{0.5em}
        \begin{subfigure}
                \centering
    \includegraphics[width=\linewidth]{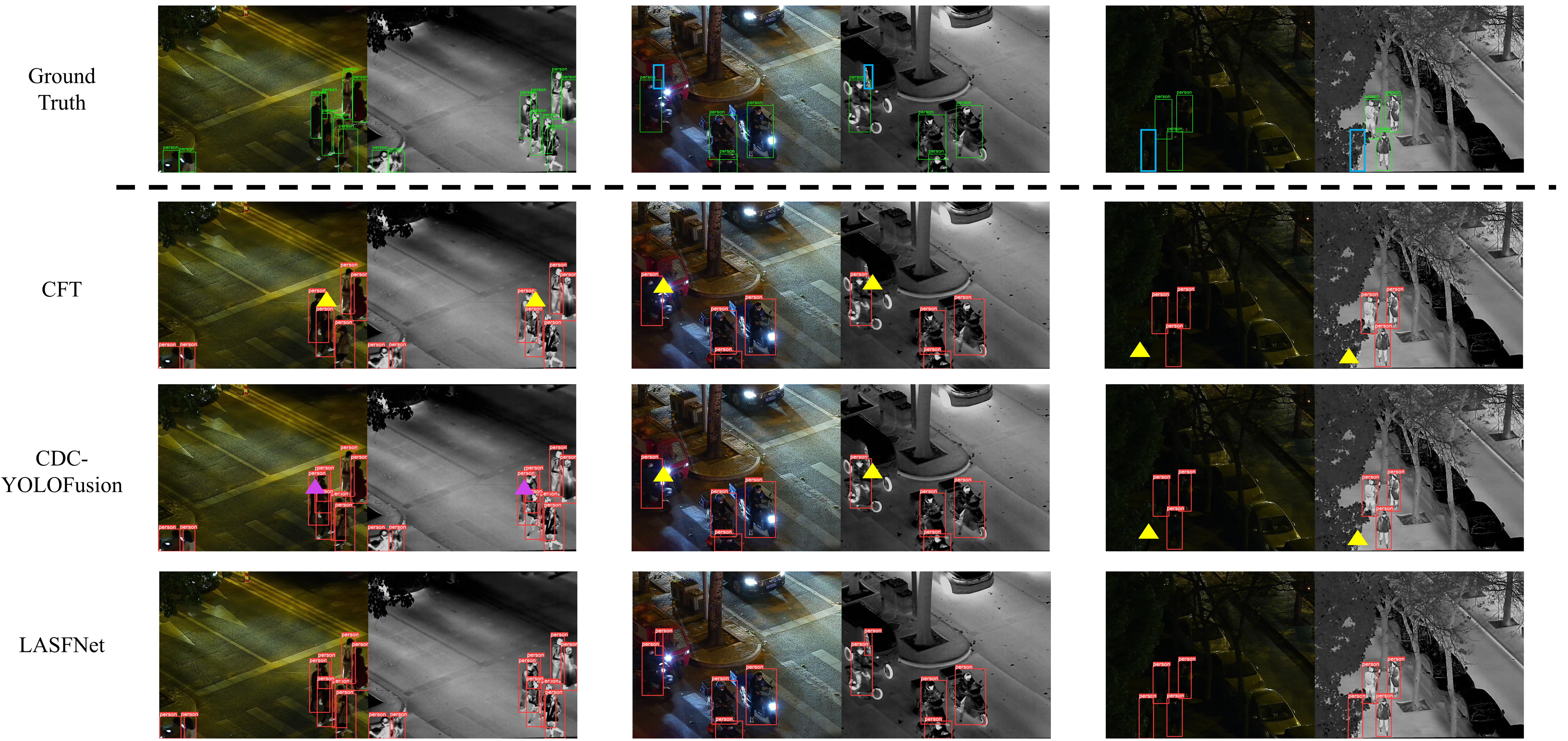}
    \caption{Three examples on LLVIP dataset to illustrate the detection results, where the purple triangles represent false positive objects, and the yellow triangles indicate the missing objects. In addition, the blue boxes highlight human annotation omissions in the ground truth.}
    \label{fig7}
        \end{subfigure}
\end{figure*}
In this section, we conduct a comparative analysis of our method against several advanced approaches on three different datasets. The comparative experimental results for the DroneVehicle, LLVIP, and VTUAV\_det datasets are presented in Tables \ref{tab:my_label4} to \ref{tab:my_label6}, respectively. Our method consistently demonstrates significant advantages in terms of both lightweight design and high accuracy across all three datasets. Although the parameter count of our model is slightly higher than that of the two lightweight models, SuperYOLO and GHOST, it offers a substantial reduction in computational complexity. Furthermore, our method achieves a notable improvement of \begin{table}[H]
    \centering
\caption{Performance Comparison Between LASFNet and Several State-of-the-art Approaches on the Dronevehicle Dataset.}
\label{tab:my_label4}
    \resizebox{0.5\textwidth}{!}{
    \begin{tabular}{lcccc}
    \toprule
         Method&    mAP50&  mAP&  Params(M)
& GFLOPs
\\
    \midrule
        (2021) CFT\cite{qingyun2021cross}&     0.849&  0.628&  44.4& 35.6
         \\[-0.2cm]\\
        (2023 TGRS) SuperYOLO\cite{Zhang2023}&     0.814&  0.595&  \textbf{4.8}& 55.9\\
         \\[-0.2cm]
        (2023 TGRS) GHOST\cite{zhang2023guided}&     0.814&  0.594&  \textbf{4.8}& 54.7\\
         \\[-0.2cm]
        (2023 RS) ADCNet\cite{he2023misaligned}&     0.776&  0.547&  25.9& 40.2\\
         \\[-0.2cm]
        (2024 PR) ICAFusion\cite{shen2024icafusion}&     0.851&  0.621&  20.1& 29.7\\
         \\[-0.2cm]
        (2024 CVPR) GM-DETR\cite{xiao2024gm}&    0.815&  0.559&  70.0& 176.0\\
         \\[-0.2cm]
        (2024 TIV) CDC-YOLOFusion\cite{wang2024cdc}&    0.848&  0.633&  33.3& 27.8\\
         \\[-0.2cm]
         LASFNet(Ours)&     \textbf{0.857}&  \textbf{0.647}&  7.7& \textbf{26.6}\\
    \bottomrule
    \end{tabular}
    }
\end{table}
\vspace{0.5em}
\noindent2-5 percentage points in mean Average Precision (mAP) compared to these two lightweight approaches across all three datasets. When compared to other methods, our LASFNet achieves up to 90\% reduction in parameters and 85\% reduction in computational cost, while simultaneously improving the detection accuracy (mAP) by 1\% to 3\% . The improvements in accuracy and efficiency can be attributed to three key factors. First, we have introduced a novel and efficient lightweight fusion detection baseline. Secondly, the lightweight ASFF module adaptively enhances the richness and prominence of fused features through attention guidance, resulting in comprehensive and effective feature fusion. Lastly, the FATM module enhances the feature representation at the neck of the network in a lightweight manner, increasing the network's focus on multi-scale features, reducing information loss, and improving the effectiveness of feature representation and fusion. Consequently, our approach achieves a favorable trade-off between efficiency and accuracy. 
\begin{table}[]
    \centering
\caption{Performance Comparison Between LASFNet and Several State-of-the-art Approaches on the LLVIP Dataset.}
\label{tab:my_label5}
    \resizebox{0.5\textwidth}{!}{
    \begin{tabular}{lcccc}
    \toprule
         Method&    mAP50&  mAP&  Params(M)
& GFLOPs
\\
    \midrule
        (2021) CFT\cite{qingyun2021cross}&     0.966&  0.630&  44.4& 35.6
         \\[-0.2cm]\\
        (2023 TGRS) SuperYOLO\cite{Zhang2023}&     0.940&  0.582&  \textbf{4.8}& 55.9\\
         \\[-0.2cm]
        (2023 TGRS) GHOST\cite{zhang2023guided}&     0.939&  0.586&  \textbf{4.8}& 54.7\\
         \\[-0.2cm]
        (2023 RS) ADCNet\cite{he2023misaligned}&     0.954&  0.604&  25.9& 40.2\\
         \\[-0.2cm]
        (2024 PR) ICAFusion\cite{shen2024icafusion}&     0.978&  0.630&  20.1& 29.7\\
         \\[-0.2cm]
        (2024 CVPR) GM-DETR\cite{xiao2024gm}&    0.971&  0.670&  70.0& 176.0\\
         \\[-0.2cm]
        (2024 TIV) CDC-YOLOFusion\cite{wang2024cdc}&   \textbf{0.979}&  0.665&  33.3& 27.8\\
         \\[-0.2cm]
         (2024) MDLNet-X\cite{wang2024unveiling}&    0.954&  0.627&  189.4& 1038.2\\
         \\[-0.2cm]
         (2024) Fusion-Mamba\cite{dong2024fusion}&    0.968&  0.628&  287.6&  -\\
         \\[-0.2cm]
         LASFNet(Ours)&     0.974&  \textbf{0.676}&  7.7& \textbf{26.6}\\
    \bottomrule
    \end{tabular}
    }
\end{table}
\begin{table}[!h]
    \centering
\caption{Performance Comparison Between LASFNet and Several State-of-the-art Approaches on the VTUAV\_det Dataset.}
\label{tab:my_label6}
    \resizebox{0.5\textwidth}{!}{
    \begin{tabular}{lcccccccccc}
    \toprule
         Method&    mAP50&  mAP&  Params(M)
& GFLOPs
\\
    \midrule
       (2021) CFT\cite{qingyun2021cross}&     0.760&  0.343&  44.4& 35.6
         \\[-0.2cm]\\
       (2023 TGRS) SuperYOLO\cite{Zhang2023}&     0.773&  0.345&  \textbf{4.8}& 55.9\\
         \\[-0.2cm]
        (2023 TGRS) GHOST\cite{zhang2023guided}&     0.477&  0.168&  \textbf{4.8}& 54.7\\
         \\[-0.2cm]
        (2023 RS) ADCNet\cite{he2023misaligned}&     0.700&  0.311&  25.9& 40.2\\
         \\[-0.2cm]
        (2024 PR) ICAFusion\cite{shen2024icafusion}&     0.762&  0.337&  20.1& 29.7\\
         \\[-0.2cm]
        (2024 CVPR) GM-DETR\cite{xiao2024gm}&    0.756&  0.350&  70.0& 176.0\\
         \\[-0.2cm]
       (2024 TIV)  CDC-YOLOFusion\cite{wang2024cdc}&    0.788&  0.357&  33.3& 27.8\\
         \\[-0.2cm]
       (2024 ISPRS) QFDet\cite{zhang2023drone}&    0.755&  0.331&  60.3& 242.8\\
         \\[-0.2cm]
        (2024 TGRS) AODet\cite{gui2024aodet}&    0.770&  0.344&  -& -\\
         \\[-0.2cm]
         LASFNet(Ours)&     \textbf{0.797}&  \textbf{0.378}&  7.7& \textbf{26.6}\\
    \bottomrule
    \end{tabular}
    }
\end{table}

\vspace{0.5em}
\begin{figure*}[]
\centering
\includegraphics[width=\linewidth]{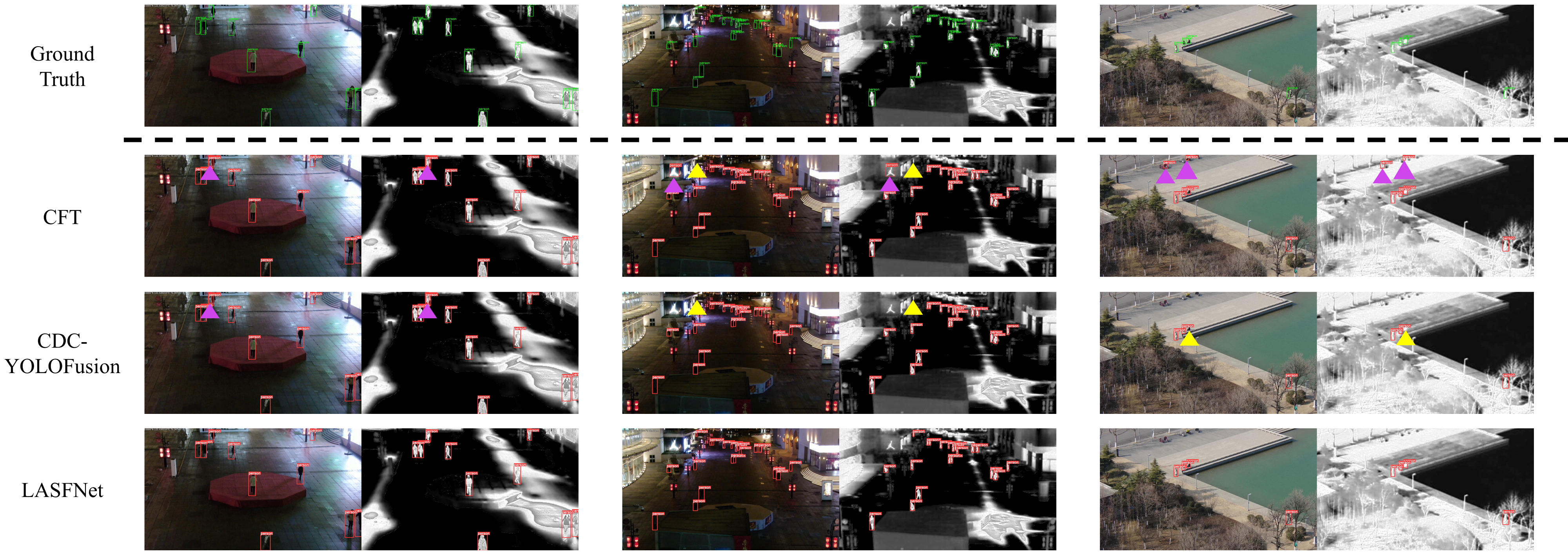}
\caption{Three examples on VTUAV\_det dataset to illustrate the detection results, where the purple triangles represent false positive objects, and the yellow triangles indicate the missing objects.  }
\label{fig8}
\end{figure*}
\begin{table}[!h]
    \centering
\caption{Model Complexity And Testing Time Comparison Among Several Multimodal Detection Approaches Tested on the LLVIP Dataset.
}
\label{tab:my_label7}
    \resizebox{0.5\textwidth}{!}{
    \begin{tabular}{lcccc}
    \toprule
         Method&     Params(M) & GFLOPs &Testing time(ms)
\\
    \midrule
        (2021) CFT\cite{qingyun2021cross}&      44.4& 35.6 &86.9
         \\[-0.2cm]\\
        (2023 TGRS) SuperYOLO\cite{Zhang2023}&      \textbf{4.8}& 55.9 &94.9\\
         \\[-0.2cm]
        (2023 TGRS) GHOST\cite{zhang2023guided}&       \textbf{4.8}& 54.7 &108.1\\
         \\[-0.2cm]
        (2023 RS) ADCNet\cite{he2023misaligned}&      25.9& 40.2 &68.4\\
         \\[-0.2cm]
        (2024 PR) ICAFusion\cite{shen2024icafusion}&       20.1& 29.7 &55.8\\
         \\[-0.2cm]
        (2024 CVPR) GM-DETR\cite{xiao2024gm}&      70.0& 176.0 &90.6\\
         \\[-0.2cm]
        (2024 TIV) CDC-YOLOFusion\cite{wang2024cdc}&     33.3& 27.8 &94.4\\
         \\[-0.2cm]
         LASFNet(Ours)&       7.7& \textbf{26.6} &\textbf{49.8} \\
    \bottomrule
    \end{tabular}
    }
\end{table}
\mbox{Table \ref{tab:my_label7}} compares the performance of various multimodal methods in terms of model parameters, computational complexity, and inference time. Our approach employs the single-fusion strategy combined with the self-modulation mechanism, achieving simultaneous improvements in model lightweighting and inference speed. Compared to existing lightweight strategies and advanced methods, our approach offers faster inference speeds with fewer parameters and reduced computational requirements, enabling more efficient detection.

Figures \ref{fig6} to \ref{fig8} present the visual detection results of CFT, CDC-YOLOFusion, and our method across three datasets. In each figure, the first row shows the ground truth labels, followed by the detection results from the three methods. Figure \ref{fig6} illustrates the detection results on the DroneVehicle dataset. Leveraging the feature modulation block (FMB) in the ASFF module, which adaptively adjusts features from different modalities, our method is able to accurately detect objects hidden in darkness while avoiding false positives caused by thermal vehicle artifacts. In contrast, both CFT and CDC-YOLOFusion exhibit some missed detections and false alarms. Figure \ref{fig7} provides the detection results on the LLVIP dataset. Our method, LASFNet, demonstrates superior performance in challenging scenarios involving dense objects and severe occlusions. This improvement is attributed to the efficient three-stage progressive complementary fusion within the ASFF module, which generates fused features with comprehensive and rich information, enabling accurate object feature coverage. In comparison, CFT and CDC-YOLOFusion show missed detections and false alarms under these challenging conditions. Figure \ref{fig8} highlights the exceptional detection capability of LASFNet in the presence of multi-scale variations and alignment issues. Compared to the other two methods, which suffer from missed detections and false positives, LASFNet benefits from the FATM module, which places strong emphasis on the fused features during transmission and facilitates the integration of multi-scale features, thereby enhancing overall detection performance.

\section{Conclusions}\label{sec:conclusions}
This paper presents a lightweight self-modulation feature fusion network, LASFNet, for multimodal object detection. LASFNet introduces a novel single-pass feature fusion approach that achieves efficient detection performance while maintaining a compact model. From a technical perspective, we design a self-modulation feature fusion module that sequentially applies attention fusion, feature modulation fusion, and channel shuffle fusion to the two input modality features. This effective design enables the network to progressively and adaptively adjust and enhance the fused features, thereby providing robust support for multimodal detection. Additionally, a feature attention transformation module is incorporated at the neck of the network to enhance feature attention and improve the fusion of multi-scale features, reducing information loss. Extensive experiments conducted on three benchmark datasets demonstrate that LASFNet is both lightweight and effective, achieving a favorable efficiency-accuracy trade-off. 

\bibliographystyle{ieeetr}
\bibliography{ref}
\end{document}